\documentclass[10pt,twocolumn,letterpaper]{article}

\usepackage[pagenumbers]{iccv} 

\definecolor{iccvblue}{rgb}{0.21,0.49,0.74}
\usepackage[pagebackref,breaklinks,colorlinks,allcolors=iccvblue]{hyperref}

\usepackage{multirow}
\usepackage{xcolor}
\usepackage{pifont}

\newcommand{\cmark}{\textcolor{green}{\ding{51}}} 
\newcommand{\xmark}{\textcolor{red}{\ding{55}}}   
\newcommand{\methodname}{SocialGen} 

\def\paperID{11847} 
\def\confName{ICCV}
\def\confYear{2025}

\def\modelnamenp{SocialGen}
\def\modelname{SocialGen}

\title{\modelnamenp: Modeling Multi-Human Social Interaction with Language Models}

\author{
    Heng Yu$^{*}$ \qquad
    Juze Zhang$^{*}$ \qquad
    Changan Chen\qquad
    Tiange Xiang\qquad
    Yusu Fang\\
    Juan Carlos Niebles\qquad
    Ehsan Adeli\\[2mm]
    Stanford University\\
}
\begin{document}


\twocolumn[{%
    \renewcommand\twocolumn[1][]{#1}%
    \maketitle
    \vspace{-0.7cm}  
    \begin{center}
        \centering
        \includegraphics[width=1.0\textwidth]{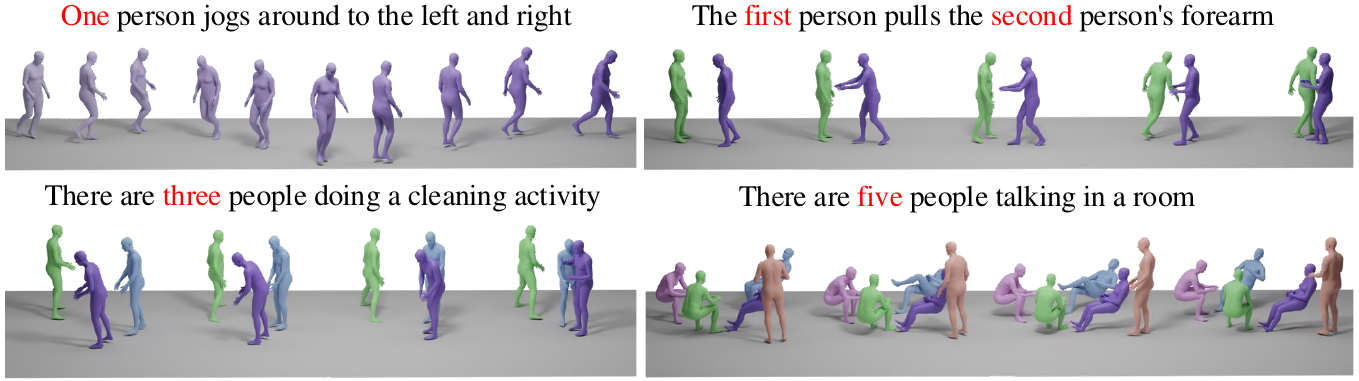}
        \vspace{-20pt}
        \captionof{figure}{SocialGen generates social interaction motion for varying numbers of individuals while also supporting diverse motion-language tasks such as captioning, completion, and reaction generation.}
        \label{fig:teaser}
    \end{center}
    \vspace{0.0cm}  
}]


\begingroup
\renewcommand\thefootnote{}\footnote{$^*$indicates equal contribution}\addtocounter{footnote}{-1}
\endgroup

\begin{abstract}
Human interactions in everyday life are inherently social, involving engagements with diverse individuals across various contexts. Modeling these social interactions is fundamental to a wide range of real-world applications.
In this paper, we introduce \methodname, the first unified motion-language model capable of modeling interaction behaviors among varying numbers of individuals, to address this crucial yet challenging problem. Unlike prior methods that are limited to two-person interactions, we propose a novel social motion representation that supports tokenizing the motions of an arbitrary number of individuals and aligning them with the language space. This alignment enables the model to leverage rich, pretrained linguistic knowledge to better understand and reason about human social behaviors. To tackle the challenges of data scarcity, we curate a comprehensive multi-human interaction dataset, \textbf{SocialX}, enriched with textual annotations. 
Leveraging this dataset, we establish the first comprehensive benchmark for multi-human interaction tasks. Our method achieves state-of-the-art performance across motion-language tasks, setting a new standard for multi-human interaction modeling.
Project page:
\href{https://socialgenx.github.io/}{\small\texttt{socialgenx.github.io}}
\end{abstract}    
\section{Introduction}
\label{sec:intro}

Modeling human behavior in social interactions are pivotal for transformative applications such as VR, gaming, healthcare, and robotics. In the entertainment field, realistic multi-human interactions enhance immersion. In healthcare, understanding patient-physician dynamics improves telemedicine and therapy. In robotics, socially aware interaction modeling enables seamless collaboration in human-centered settings. These applications highlight the significance of modeling multi-human social scenarios.


Recent years have witnessed remarkable progress in human motion understanding~\cite{jiang2023motiongpt, chen2024language, wu2024motionllm} and generation~\cite{plappert2018learning, lin2018generating, ghosh2021synthesis}. However, most existing work focuses on single-person~\cite{jiang2023motiongpt, chen2024language, plappert2018learning} or dyadic (two-person) interactions~\cite{liang2024intergen, javed2024intermask}, often relying on hard-coded interaction models that cannot generalize to scenarios with arbitrary numbers of participants. Real-world interactions, by contrast, are dynamic and frequently involve multiple individuals with complex dependencies. Addressing these challenges necessitates a scalable approach capable of modeling social interactions for any number of humans, supporting both understanding and generation.

To overcome the limitations in more general interaction modeling, we propose a novel motion-language framework that generates motions for any number of individuals and enables various motion-related tasks. 
The key to our approach is a novel motion representation \textbf{XH3D}, specifically designed to encode and decode complex multi-human interactions effectively. We tokenize motion, spatial, and text data using modality-specific tokenizers, creating a unified vocabulary that integrates text, individuals, and relative spatial configurations. These tokens are processed by a motion-language model, which outputs sequences of modality tokens. Our framework includes a pre-training stage to spatially and temporally capture human interactions, aligning modalities in the latent space, followed by post-training with diverse task instructions to enhance generalization. This unified approach enables motion generation for any number of individuals and supports diverse tasks like captioning, completion, and reaction generation.


While several motion datasets for multi-human social interactions exist~\cite{zhang20204d,van2011umpm,von2018recovering}, they lack language descriptions, limiting their utility for tasks involving language input or output. To address this, we introduce \textbf{SocialX}, a curated dataset enriched with detailed captions generated using GPT-4o. By employing a multi-step prompting scheme, we ensure accurate and faithful descriptions of multi-human social interactions, enabling benchmarking for generation and other motion-language tasks.

To validate our model, we first evaluate the effectiveness of our motion tokenizer on InterHuman and InterX datasets, demonstrating superior reconstruction and better capture of inter-human dynamics compared to existing representations. We further benchmark our model on SocialX, where it outperforms state-of-the-art methods across multiple metrics, generating smoother and more natural human motions. Additionally, our method supports various other motion-related tasks.
In a nutshell:
\begin{itemize}
    \item We propose a novel approach to modeling multi-human social interactions by leveraging language models while also supporting various downstream motion-related tasks.
    \item We introduce a novel motion representation for scenarios involving varying numbers of participants, enabling scalable and effective modeling of social interactions.
    \item We develop \textbf{SocialX}, a curated dataset paired with a benchmark, encompassing diverse scenarios with human pose data and language descriptions, setting a new standard for evaluating multi-human interaction models.
\end{itemize}
We believe that our framework and benchmark will provide a strong foundation for future research in multi-human interaction modeling.

\section{Related Work}
\label{sec:formatting}

\noindent\textbf{Human Motion Generation and Understanding:}
Generating and understanding human motion are key to human-centric AI. Recent works explore realistic and diverse motion generation using multimodal inputs, including textual descriptions~\cite{plappert2018learning, lin2018generating, ghosh2021synthesis, ahuja2019language2pose, tevet2023human, zhang2023generating, guo2022generating, guo2022tm2t, zhang2024motiondiffuse, petrovich2022temos, kim2023flame, yi2024generating, zhang2023remodiffuse, barquero2024seamless}
actions~\cite{petrovich2021action, guo2020action2motion, chen2023executing}, music or audio~\cite{chen2024language, dabral2023mofusion, tseng2023edge, zhou2023ude, li2024dance, siyao2022bailando, ng2022learning, ng2024audio, le2023music, ma2022pretrained, zhao2023taming}, 
incomplete motion~\cite{butepage2017deep, chen2023humanmac, martinez2017human, yuan2020dlow, zhang2021we, ma2022multi},
and other control signals~\cite{goel2024iterative, huang2024controllable, jiang2024motionchain, petrovich2024multi, wan2023tlcontrol, xie2023omnicontrol, tessler2024maskedmimic}
Among these, text-to-motion generation is gaining attention for its intuitive and naturally use-friendly input. Diffusion-based models~\cite{zhou2024emdm, zhang2024motiondiffuse, shafir2023human, tevet2023human, wang2023fg, kong2023priority, zhang2023finemogen, zhang2023remodiffuse, mandelli2024generation, ma2022pretrained, chen2023humanmac}
and vector-quantized approaches~\cite{van2017neural,guo2022tm2t, guo2024momask, zhang2023generating, lucas2022posegpt, pinyoanuntapong2024mmm}
have demonstrated promising results in generating isolated human motion. However, they struggle to capture broader motion patterns, particularly in complex social interactions. Recent works~\cite{gong2023tm2d, jiang2023motiongpt, zhang2024motiongpt, zhou2023ude, luo2024m, sun2024coma, liu2024emage, cai2024digital, Wu2024MotionAgentAC, zhang2024large, feng2024chatpose, lin2024chathuman, li2024unimotion} unify tasks or modalities, but lack multi-human handling. In contrast, our work delves into a broader set of scenarios, explicitly tackling social group motion generation.
\noindent\textbf{Human-Human Interaction Modeling:}
Most motion generation methods focus on single humans, while interactions add complexity by requiring inter-individual and spatio-temporal modeling.
Recent works on human-human interaction modeling mainly focuse on two-person scenarios, addressing (1) \textit{reaction synthesis}, where a reactor responds to an actor’s motion~\cite{chopin2023interaction, liu2023interactive, xu2024regennet, ghosh2024remos}, and (2) \textit{interaction generation}, where all individuals’ motions are generated simultaneously~\cite{cai2024digital, shafir2023human, tanaka2023role, liang2024intergen}. Some studies extend to multi-person settings~\cite{shan2024towards, fan2024freemotion}, but they focus on motion generation and lack language-related comprehensive capabilities, limiting generalization to larger groups and open-domain prompts (see Tab.~\ref{tab:review} in the Appendix).

\noindent\textbf{Multimodal Language Models:}
Large language models (LLMs) such as T5~\cite{raffel2020exploring}, Flan-T5~\cite{chung2024scaling}, and LLaMA~\cite{touvron2023llama}, along with others~\cite{chiang2023vicuna, devlin2018bert, dai1901transformer, brown2020language}, have demonstrated exceptional capabilities in comprehension and generation tasks. Recently, researchers have extended LLMs to multimodal domains, integrating inputs from images~\cite{girdhar2023imagebind, huang2023language, li2022blip}, videos~\cite{xu2021videoclip, liu2024world}, audio~\cite{girdhar2023imagebind, guzhov2022audioclip}, and other modalities~\cite{chen2023x, han2024onellm}, paving the way for multimodal large language models (MLLMs)~\cite{zhan2024anygpt, wu2023next}. For 3D motion, MotionGPT~\cite{jiang2023motiongpt} frames single-human motions as a language modeling task, which was extended by~\cite{chen2024language, luo2024m} to include more modalities.

Building on these advancements, our work introduces the first motion-language model tailored for multi-human social interaction modeling, enabling the generation and other tasks among multiple individuals.

\noindent\textbf{Human Motion Datasets:}
Datasets are fundamental for advancing motion-related synthesis tasks. Single-human datasets~\cite{liu2019ntu, punnakkal2021babel, guo2022generating, plappert2016kit, lin2023motion} have been instrumental in developing motion-related models but face challenges when generalizing to multi-human scenarios. While datasets like InterGen~\cite{liang2024intergen}, InterX~\cite{xu2024inter}, and InterAct~\cite{huang2024interact} provide two-person interactions with text descriptions, they are limited in addressing broader multi-human contexts.

Several multi-human motion datasets, including 4DAssociation \cite{zhang20204d}, UMPM~\cite{van2011umpm}, 3DPW~\cite{von2018recovering}, MuPoTS-3D~\cite{mehta2018single}, ExPI~\cite{guo2022multi}, CMU-Panoptic~\cite{joo2019towards}, You2Me~\cite{ng2020you2me}, BEDLAM~\cite{black2023bedlam}, and HOI-M$^3$~\cite{zhang2024hoi}, have been introduced. However, many of these datasets lack text descriptions, limiting their utility in multimodal understanding. 
Recent datasets like LAION-Pose and WebVid-Motion~\cite{shan2024towards} include pose/motion and text pairs extracted from in-the-wild images and videos~\cite{schuhmann2021laion}, but their annotations and descriptions are constrained by automated pipelines.

In this paper, we introduce \textit{SocialX}, a novel dataset with accurate SMPL-compatible~\cite{loper2023smpl} annotations and high-quality text descriptions. \textit{SocialX} captures interactions involving varying numbers of individuals and establishes a benchmark for multi-human motion modeling, offering valuable resources for advancing the field.

\section{Method}

\begin{figure*}[htp]
  \centering
  \includegraphics[width=\textwidth]{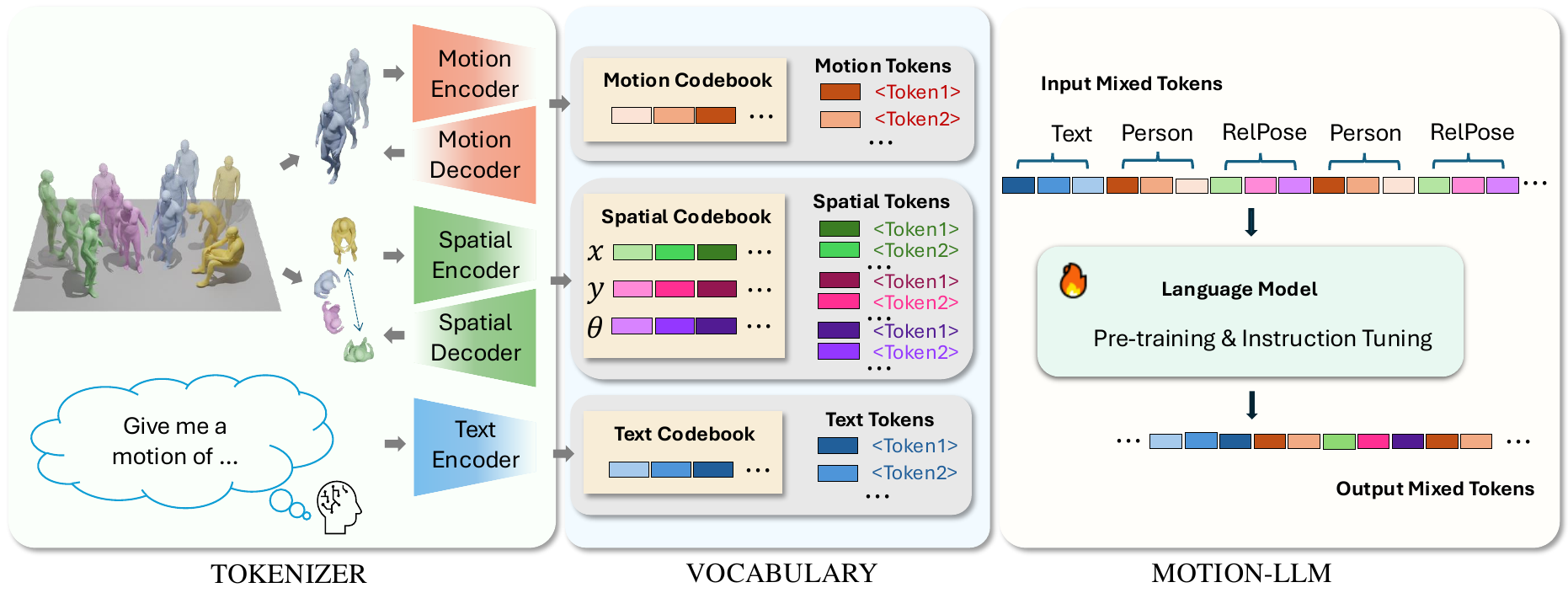}
  \caption{Overview of \modelname Framework. \modelname unifies motion generation, understanding, and other motion-language tasks for multi-human social interactions. The tokenizer encodes motion, spatial, and text features into tokens via respective codebooks, forming a unified vocabulary of motion, spatial, and text Tokens. These tokens, combined into mixed input tokens representing text, individuals, and relative spatial configurations, are processed by the Motion-LLM, which leverages pre-training and instruction tuning to generate mixed tokens, enabling versatile modeling of complex social interactions.}
  \label{pipeline}
  \vspace{-0.3cm}
\end{figure*}

We introduce \modelname, a unified motion-language framework that models interaction motions for varying group sizes by aligning motion and language spaces.
As shown in Figure~\ref{pipeline}, SocialGen consists of two key components, similar to~\cite{jiang2023motiongpt}: a motion tokenizer that discretizes motion data and a motion-aware language model that aligns motion and language tokens. This alignment enables motion modeling using pre-trained language models and textual descriptions.
\subsection{Motion Representation}
\label{sec:mot_rep}
\noindent\textbf{Existing Representations}  
Existing motion representations primarily cater to either single-human or two-human.  

\begin{itemize}
    \item \textbf{H3D}: Designed for single-human motions, H3D~\cite{guo2022generating} provides a canonical representation by transforming joint positions and velocities into the root frame. A pose is defined as $(r^a, r^x, r^z, r^y, \mathbf{j}^p, \mathbf{j}^v, \mathbf{j}^r, \mathbf{c}^f)$, where $r^a \in \mathbb{R}$ is the root angular velocity along the Y-axis,  
    $(r^x, r^z) \in \mathbb{R}^2$ are root linear velocities on the XZ-plane, and  
    $r^y \in \mathbb{R}$ is root height.  
    The local joint positions, velocities, and rotations are given by  
    $\mathbf{j}^p \in \mathbb{R}^{3(j-1)}$,  
    $\mathbf{j}^v \in \mathbb{R}^{3j}$, and  
    $\mathbf{j}^r \in \mathbb{R}^{6(j-1)}$,  
    with $j$ denoting the number of joints.  
    Binary foot-ground contact features $\mathbf{c}^f \in \mathbb{R}^4$ are obtained by thresholding heel and toe joint velocities~\cite{guo2022generating}. With a 6D rotation format~\cite{zhou2019continuity} and the SMPL skeleton~\cite{loper2023smpl}, the pose has 263 dimensions. While effective for single-human motion, H3D lacks the ability to model multi-human interactions.
    
    \item \textbf{IH}: The IH format~\cite{liang2024intergen} models multi-human interactions by encoding absolute global joint trajectories in the world frame. A pose is represented as $(\mathbf{j}^p_g, \mathbf{j}^v_g, \mathbf{j}^r, \mathbf{c}^f)$, where $\mathbf{j}^p_g\in \mathbb{R}^{3j}$ and $\mathbf{j}^v_g\in \mathbb{R}^{3j}$ are global joint positions and velocities, and $\mathbf{j}^r$ and $\mathbf{c}^f$ match the H3D definitions. With the SMPL skeleton, the pose dimension is 262. However, IH suffers from reconstruction artifacts during motion tokenization due to its high-dimensional global position.

    \item \textbf{IX}: Directly using SMPL~\cite{loper2023smpl} parameters, IX is a straightforward representation employed by datasets like InterX~\cite{xu2024inter}. While conceptually simple, it faces similar challenges as IH, such as jittering and artifacts during tokenization, owing to its high-dimensional encoding.
\end{itemize}

\noindent\textbf{Proposed XH3D} 
Although the IH and IX formats are capable of representing multi-human scenarios, they struggle to achieve high-quality reconstruction during motion tokenization. This is because the token codebook typically consists of very few tokens (usuallly hundreds), while the root global position and 6D continuous SMPL representation are both high-dimensional. Consequently, IH and IX formats often suffer from jittering and artifacts after motion tokenization, as demonstrated in Section~\ref{sec:exp}.
To address this issue, we propose XH3D, an extension of the H3D format for multi-human scenarios. For each individual, XH3D adopts the same tuple $(r^a, r^x, r^z, r^y, \mathbf{j}^p, \mathbf{j}^v, \mathbf{j}^r, \mathbf{c}^f)$ used in H3D as the motion representation. Additionally, to capture relative spatial relationships between humans, we introduce a tuple $(s^x, s^z, s^\theta)$ representing the first-frame relative pose between two motions. Here, $(s^x, s^z) \in \mathbb{R}^2$ denotes the relative position on the root XZ-plane, and $s^\theta \in \mathbb{R}$ represents the relative orientation. For multiple humans, one is randomly chosen as the reference, and the relative pose of each remaining human is computed accordingly. This randomness also serves as data augmentation.
We assert that XH3D is scalable to any number of individuals and achieves superior tokenization performance compared to existing formats, as demonstrated in Table~\ref{tb:token_cmpr}.

\subsection{Motion Tokenizer}
To represent motion as discrete tokens, we employ a 3D human motion tokenizer based on the Vector Quantized Variational Autoencoder (VQVAE) framework \cite{van2017neural, gong2023tm2d,jiang2023motiongpt,zhou2023ude,luo2024m, zhang2024motiongpt}. The motion tokenizer consists of an encoder $\mathcal{E}$, a decoder $\mathcal{D}$, and a learnable codebook $\mathcal{Z} = \{z_1, z_2, \dots, z_K\}$ containing $K$ discrete latent embedding vectors of dimension $d$. The encoder extracts latent motion features that are quantized into discrete tokens using the nearest entries from the codebook. The decoder reconstructs the motion sequence from these quantized tokens, enabling a compact and discrete representation suitable for downstream tasks. We independently tokenize each individual using VQVAE, following a similar approach to \cite{jiang2023motiongpt}.

The training process optimizes a combination of reconstruction, embedding, and commitment losses to ensure high-quality motion representation following \cite{jiang2023motiongpt, zhang2024motiongpt}. The loss function is defined as:
\[
\mathcal{L}_\text{VQ} = \underbrace{\|\hat{\mathbf{m}} - \mathbf{m}\|_1}_{\mathcal{L}_r} + \underbrace{\|\text{sg}[\mathbf{z}] - \mathbf{e}\|_2^2}_{\mathcal{L}_e} + \underbrace{\beta \|\mathbf{z} - \text{sg}[\mathbf{e}]\|_2^2}_{\mathcal{L}_c},
\]
where $\mathbf{m}$ is the original motion sequence, $\hat{\mathbf{m}}$ is the reconstructed motion, $\mathbf{z}$ are the latent embeddings, $\mathbf{e}$ are the codebook entries, $\text{sg}[\cdot]$ denotes the stop-gradient operation, and $\beta$ is a weighting factor for the commitment loss.

Additionally, we follow \cite{zhang2023generating} to incorporate L1 smooth loss for reconstruction and velocity regularization to improve motion generation quality. Exponential moving average (EMA) is applied to update the codebook during training \cite{razavi2019generating}, ensuring stable and efficient codebook utilization. This approach enables efficient motion tokenization, facilitating the representation of motion as discrete tokens for seamless integration with language models and improved performance in motion-related tasks.

For the relative root pose tuple \((s^x, s^z, s^\theta)\), we apply uniform binning to discretize each component into \(K\) tokens by dividing its range—determined by dataset-wide extrema—into equal intervals. While simple and effective, alternative methods such as non-uniform bins, larger \(K\), or vector quantization could improve performance. However, our focus is on the framework of separately tokenizing the relative pose to handle arbitrary numbers of individuals, while the optimal tokenization strategy remains an open problem beyond this study.


\subsection{Motion-Language Model}
To seamlessly integrate text and motion modeling, we employ a motion-aware language model built on a pre-trained large language model (LLM) such as T5~\cite{raffel2020exploring}. Our approach leverages the proposed motion tokenizer to map a motion sequence $m^{1:M}$ into a sequence of discrete motion tokens $z^{1:L}$, enabling the unified modeling of motion and text within a shared vocabulary~\cite{jiang2023motiongpt, kudo2018sentencepiece, ouyang2022training, raffel2020exploring}. Each motion token $z_i$ corresponds to an index in the motion codebook, including tokens for relative root poses, ensuring compatibility with the text token vocabulary. To achieve this, we extend the original text token vocabulary $V_t$ with motion tokens $V_m = \{\langle \text{Motion}_i \rangle \}_{i=1}^K$ and relative root pose tokens $V_x = \{\langle \text{x}_i \rangle \}_{i=1}^K$, $V_z = \{\langle \text{z}_i \rangle \}_{i=1}^K$, and $V_\theta = \{\langle \mathrm{\theta}_i \rangle \}_{i=1}^K$, forming a unified vocabulary $V = \{ V_t, V_m, V_x, V_z, V_\theta \}$. This unified representation allows both input and output sequences to seamlessly incorporate natural language and motion tokens (including relative root pose tokens).


Special tokens, such as $\langle \text{Motion}_S \rangle$ and $\langle \text{Motion}_E \rangle$, mark the start and end of a motion sequence. For multiple humans, the tokens $\langle \text{x}_i \rangle$, $\langle \text{z}_i \rangle$, and $\langle \mathrm{\theta}_i \rangle$ serve both as separators between different individuals and as representations of their relative poses. The general format for encoding motion tokens in multi-human scenarios is:
\[
\begin{aligned}
    &\langle \text{Motion}_S \rangle \ldots \langle \text{Motion}_{i-1} \rangle
    \langle \text{x}_i \rangle \langle \text{z}_i \rangle \langle \mathrm{\theta}_i \rangle 
    \langle \text{Motion}_i \rangle \\
    &\langle \text{x}_{i+1} \rangle \langle \text{z}_{i+1} \rangle 
    \langle \mathrm{\theta}_{i+1} \rangle \langle \text{Motion}_{i+1} \rangle \ldots 
    \langle \text{Motion}_E \rangle
\end{aligned}
\]

where each triplet $\langle \text{x}_i \rangle \langle \text{z}_i \rangle \langle \mathrm{\theta}_i \rangle$ uniquely identifies a human and encodes their relative pose. This structure ensures scalability to any number of humans within a scene. These special tokens significantly enhance the model’s ability to model multi-human interactions. Unless otherwise stated, motion tokens in the following text include both individual motion and relative pose information.

During training, the model learns to represent the text-motion relationship as an autoregressive sequence generation task. Given an input sequence $X_s = \{ x_s^i \}_{i=1}^N$ consisting of text and motion tokens of length $N$ and an output sequence $X_t = \{ x_t^i \}_{i=1}^L$ of length $L$, the model predicts the probability of each token conditioned on the preceding tokens:
\[
p_\theta(x_t | x_s) = \prod_{i} p_\theta(x_t^i | x_t^{<i}, x_s),
\]
The training objective maximizes the log-likelihood of the data distribution:
\[
\mathcal{L}_{LM} = - \sum_{i=0}^{L_t - 1} \log p_\theta(x_t^i | x_t^{<i}, x_s).
\]

At inference time, the model generates text or motion tokens iteratively in an autoregressive manner. For instance, given a text prompt describing a motion, the model predicts a sequence of motion tokens until the $\langle \text{Motion}_E \rangle$ token is generated, marking the completion of the motion sequence. This iterative approach supports flexible motion generation with variable lengths and accommodates varying numbers of humans, adapting to the specific requirements of the input description. By incorporating motion into the language modeling framework, our motion-aware language model bridges the gap between text and motion, delivering robust performance on tasks requiring the joint understanding and generation of text and motion data, particularly in scenarios involving multi-human interaction behaviors.

\subsection{Training Paradigm}
As shown in Figure~\ref{pipeline}, our training process consists of three stages: (1) Motion Tokenizer Training, which learns a motion codebook to represent motion as discrete tokens; (2) Motion-Language Alignment Pre-training, which aligns motion and text modalities using a unified vocabulary and both supervised and unsupervised objectives; and (3) Instruction Tuning, which fine-tunes the model with prompt-based instructions to enhance its performance on diverse motion-relevant tasks. \\
\noindent\textbf{Motion Tokenizer Training} 
We first train the motion tokenizer using the objective defined by $\mathcal{L}_\text{VQ}$. This process maps any human motion sequence $m^{1:M}$ into a sequence of discrete motion tokens $z^{1:L}$, enabling seamless integration with text within the unified model. Once trained, the motion tokenizer remains fixed throughout subsequent stages of the pipeline to ensure stability and consistency. For the relative pose tokenizer, uniform binning is employed, eliminating the need for additional training.

\noindent\textbf{Motion-language Alignment Pre-training} 
The T5 model~\cite{raffel2020exploring}, pre-trained and fine-tuned~\cite{chung2024scaling} on natural language datasets, is further pre-trained on our SocialX dataset, which combines text and motion data involving various numbers of people. This pre-training occurs in both unsupervised and supervised settings. To improve the model's generalization ability across diverse tasks~\cite{jiang2023motiongpt, radford2019language, devlin2018bert, raffel2020exploring, ouyang2022training}, we adopt an unsupervised learning objective~\cite{jiang2023motiongpt, raffel2020exploring}. Specifically, $15\%$ of the tokens in the input sequence $X_s$ are randomly replaced with a special sentinel token. The target sequence is then constructed by extracting the replaced token spans, delimited by sentinel tokens, with an additional sentinel token marking the end of the target sequence. Further details can be found in~\cite{jiang2023motiongpt}.

In the supervised setting, the model is trained on paired text-motion datasets through specifically designed tasks. The first task, motion-language alignment, involves generating paired data where the input is either a motion sequence or a text description, and the output is its corresponding counterpart. The second task, motion forecasting, trains the model to predict future motion sequences based on a given initial motion. The third task, reaction synthesis, involves masking the motion of one individual in a multi-human scenario, with the model tasked to generate the missing motion.

Through these unsupervised and supervised pre-training stages, the model learns to generate and understand coherent text and motion representations, effectively bridging the gap between these modalities and achieving robust performance on motion-related tasks.

\noindent\textbf{Post-training with Instruction Tuning} 
To enhance the generalization and instruction-following capabilities of our motion-language model, we construct a diverse set of tasks focusing on motion captions, human count, frame count, and motion length, leveraging our SocialX dataset. Specifically, we define 25 core motion-related tasks, including motion completion, motion generation from text, motion captioning, motion prediction, and human number estimation from motion data. For each task, we create a variety of instruction templates using GPT-4 \cite{achiam2023gpt}, resulting in a dataset containing thousands of unique task instructions. Additional examples of these templates can be found in the Supp.



\section{Dataset}
To create the benchmark for multi-person interactions, we enriched existing motion datasets with detailed text descriptions. This led to \textbf{SocialX}, the first dataset capturing diverse social interactions across varying group sizes and scenarios, including home, workplace, and recreational settings.

\subsection{Datset Sources}
Our \textbf{SocialX} dataset integrates and extends multiple existing datasets, including HumanML3D~\cite{guo2022generating}, InterHuman~\cite{liang2024intergen}, InterX~\cite{xu2024inter}, and HOI-M$^3$~\cite{zhang2024hoi}, CMU Panoptic~\cite{joo2019towards}, and MuPoTS-3D~\cite{mehta2018single}, as summarized in Table~\ref{data_sum} in the Appendix. To further enrich SocialX, we synthesize multi-human interactions by combining motions from single-person~\cite{guo2022generating} and two-person~\cite{liang2024intergen, xu2024inter} datasets while ensuring no physical conflicts.  
Although these synthesized interactions may not always form meaningful full-group scenarios, they help the model learn participant counts and sub-group interactions. SocialX also provides detailed text descriptions and a unified pose representation, making it the first large-scale dataset for multi-human interaction modeling. See the Appendix for details.

\subsection{Caption Generation}
For datasets lacking text descriptions, we use GPT-4o-mini~\cite{openai2024gpt4o} to generate content descriptions from videos. Frames are sampled at 1 fps, and videos are segmented into 20-second clips. When multi-view images (e.g., in HOI-M3) are available, they are utilized to enhance caption accuracy and reduce occlusion effects. Descriptions are generated at two levels, with manual filtering applied to ensure clean and accurate text. See the Appendix for details.

\noindent\textbf{Scene Description} We prompt GPT-4o-mini to generate detailed descriptions of human motion and interactions in the video clips. While the focus is on motion, scene-level descriptions may also include human-object interactions, appearances, and other contextual details.

\noindent\textbf{Human Interaction Description} To create captions focused purely on human interactions, we refine the scene-level descriptions using GPT-4o-mini. This process removes unrelated information, ensuring the captions concentrate solely on human motion and interactions. The result is a detailed, behavior-level description.


\subsection{Motion Data Postprocessing}
To ensure consistency across all datasets, we preprocess the motion data into a unified format compatible with our framework. For datasets that do not provide SMPL parameters, we first estimate these parameters by fitting the SMPL model~\cite{loper2023smpl} to the motion data. This process involves optimizing the pose and shape parameters to best align the motion capture data with the SMPL representation.

Once the SMPL parameters are obtained, we process the motion of each individual following the methodology introduced in HumanML3D~\cite{guo2022generating}. Specifically, we extract individual motions and normalize them to a standard reference. Additionally, as discussed in Section~\ref{sec:mot_rep}, we tokenize the relative positions between individuals to encode inter-person spatial relationships effectively. This step is crucial for capturing social interactions and enables our model to generalize across multi-human scenarios.

We utilize the validation and test sets from HumanML3D~\cite{guo2022generating}, InterHuman~\cite{liang2024intergen}, InterX~\cite{xu2024inter}, and HOI-M$^3$~\cite{zhang2024hoi}. The remaining datasets, including those with limited multi-human interaction scenarios such as MuPoTS-3D~\cite{mehta2018single} and the Haggling sequences from CMU Panoptic~\cite{joo2019towards}, are incorporated into the training set. To ensure consistency across multi-human datasets, all sequences are downsampled to 20 fps, and longer sequences are segmented into shorter clips, following the captioning preprocessing strategy described earlier. By integrating data from diverse sources, our training set captures a broad range of motion scenarios, enhancing the robustness and generalization capability of our model.
By utilizing this pipeline, all motion data are transformed into our standardized format, allowing for seamless integration and consistent processing in our framework. More details are in the Appendix.


\begin{table*}[h]
\centering
\resizebox{\textwidth}{!}{%
\begin{tabular}{cccccccccc}
\toprule
\textbf{Dataset} & \textbf{Method} & \multicolumn{3}{c}{\textbf{R Precision$\uparrow$}} & \textbf{FID$\downarrow$} & \textbf{MM Dist$\downarrow$} & \textbf{Diversity$\rightarrow$} & \textbf{MModality$\uparrow$} \\ 
\cmidrule(lr){3-5}
 & & Top 1 & Top 2 & Top 3 & & & & \\ 
\midrule
\multirow{7}{*}{SocialX} & Ground Truth & 0.542\textsuperscript{$\pm$.002} & 0.752\textsuperscript{$\pm$.002} & 0.853\textsuperscript{$\pm$.002} & 0.0002\textsuperscript{$\pm$.0001} & 3.676\textsuperscript{$\pm$.006} & 14.334\textsuperscript{$\pm$.114} & - \\

 & MDM~\cite{tevet2023human} & 0.281\textsuperscript{$\pm$.003} & 0.563\textsuperscript{$\pm$.003} & 0.672\textsuperscript{$\pm$.003} & 4.260\textsuperscript{$\pm$.006} & - & - & - \\

 & T2M~\cite{guo2022generating} & \textbf{0.409}\textsuperscript{$\pm$.003} & \underline{0.599}\textsuperscript{$\pm$.004} & \underline{0.707}\textsuperscript{$\pm$.002} & 10.694\textsuperscript{$\pm$.092} & \textbf{5.081}\textsuperscript{$\pm$.010} &  13.647\textsuperscript{$\pm$.121} & 0.824\textsuperscript{$\pm$.047} \\
 & TM2T~\cite{guo2022tm2t} & 0.390\textsuperscript{$\pm$.003} & 0.584\textsuperscript{$\pm$.003} & 0.696\textsuperscript{$\pm$.002} & 2.349\textsuperscript{$\pm$.007} & 5.479\textsuperscript{$\pm$.013} & \underline{14.109}\textsuperscript{$\pm$.130} & 2.873\textsuperscript{$\pm$.073}. \\

 & T2M-GPT~\cite{zhang2023generating} & 0.403\textsuperscript{$\pm$.003} & 0.588\textsuperscript{$\pm$.003} & 0.706\textsuperscript{$\pm$.004} & 10.163\textsuperscript{$\pm$.007} & \underline{5.097}\textsuperscript{$\pm$.005} & 13.359\textsuperscript{$\pm$.121} & 0.9186\textsuperscript{$\pm$.051}. \\
\cmidrule(lr){2-9}

 & Ours - Pre-training & 0.396\textsuperscript{$\pm$.004} & 0.593\textsuperscript{$\pm$.002} & 0.699\textsuperscript{$\pm$.003} & \underline{1.795}\textsuperscript{$\pm$.005} & 5.662\textsuperscript{$\pm$.012} & 13.898\textsuperscript{$\pm$.139} & \underline{3.028}\textsuperscript{$\pm$.077} \\ 

 & Ours - Instruction Tuning & \underline{0.403}\textsuperscript{$\pm$.003} & \textbf{0.601}\textsuperscript{$\pm$.002} & \textbf{0.717}\textsuperscript{$\pm$.002} & \textbf{1.403}\textsuperscript{$\pm$.005} & 5.389\textsuperscript{$\pm$.011} & \textbf{14.213}\textsuperscript{$\pm$.147} & \textbf{3.263}\textsuperscript{$\pm$.098} \\ 
\bottomrule
\end{tabular}%
}
\caption{\textbf{Quantitative evaluation on our SocialX test sets.} $\pm$ indicates a 95\% confidence interval and $\rightarrow$ means the closer to ground truth the better. \textbf{Bold face} indicates the best result, while \underline{underscore} refers to the second best.}
\label{quantitative}
\vspace{-0.2cm}
\end{table*}

\section{Experiment} 
\label{sec:exp}
\subsection{Evaluation Metrics}

To evaluate our model, we employ metrics assessing motion quality, diversity, multimodality, and text-motion alignment. Briefly, these include Frechet Inception Distance (FID) for motion quality, Diversity (DIV) and Multimodality (MM) for variation and distinctiveness, and R-Precision and Multi-Modal Distance (MM Dist) for text-motion alignment. Detailed descriptions of these metrics are provided in the Appendix. To evaluate the reconstruction error of VQVAE using different motion representations, we report the mean per joint position error (MPJPE) and the Procrustes Analysis MPJPE (PA-MPJPE) \cite{gower1975generalized} for global/local errors in millimeters, the mean per joint acceleration error (Accel) for temporal quality as previous work \cite{lu2023humantomato, chen2023executing, zeng2022smoothnet}.
Additional implementation details can be found in the Appendix.



\subsection{Tokenization Evaluation}
To assess the effectiveness of our proposed pose representation, we evaluate IH, IX, and our XH3D pose representations on the InterHuman and InterX datasets independently as shown in Table~\ref{tb:token_cmpr}. The results demonstrate that our XH3D representation achieves superior performance in terms of joint position accuracy, measured by MPJPE and PA-MPJPE. Importantly, it significantly outperforms other representations in temporal quality, as reflected by the Accel metric, which is crucial for ensuring motion smoothness.

\begin{table}[h]
\centering
{\small
\renewcommand{\arraystretch}{.4}
\begin{tabular}{llccc}
\toprule
\textbf{Dataset} & \textbf{Method} & \textbf{MPJPE↓} & \textbf{PAMPJPE↓} & \textbf{ACCEL↓} \\
\midrule
\multirow{3}{*}{InterHuman} 
    & IH & 100.30 & 71.94 & 26.73 \\
    & IX   & 72.52  & 49.58 & 44.00 \\
    & XH3D        & \textbf{65.42}  & \textbf{43.52} & \textbf{8.84} \\
\midrule
\multirow{3}{*}{InterX} 
    & IH & 100.32  & 51.49 & 13.89 \\ 
    & IX  & 90.19  & 44.53 & 31.85 \\
    & XH3D          & \textbf{56.36}  & \textbf{34.97} & \textbf{5.71} \\
\bottomrule
\end{tabular}
}
\caption{Comparison of MPJPE, PAMPJPE, and ACCEL across methods for InterHuman and InterX datasets. ↓ indicates lower values are better. Bold numbers indicate the best results. Our motion representation achieves the best reconstruction quality.}
\label{tb:token_cmpr}
\vspace{-0.3cm}
\end{table}


\subsection{Quantitative Comparison}
Table~\ref{quantitative} presents a quantitative comparison of our method against previous approaches. Existing evaluation models, such as those in \cite{jiang2023motiongpt, javed2024intermask}, are trained on datasets with either single-person or two-person motions using different motion representations (e.g., IH or IX). For two-person interactions, we cannot directly use evaluation models like \cite{javed2024intermask} due to representation differences. For single-person motions, our model performs similarly to MotionGPT, though single-person generation is not the focus of this paper. 

Since no existing evaluation model can assess text-to-motion generation for an arbitrary number of individuals, we train an evaluation model on SocialX following the approach in~\cite{guo2022generating}. To ensure compatibility with varying numbers of individuals, we modify the input and output dimensions of~\cite{guo2022generating} to support up to \(N\) individuals (setting \(N=5\) in our experiments) while keeping all other configurations unchanged. Our evaluation model is designed to assess motion generation across different numbers of individuals, providing a more comprehensive evaluation of a model’s ability to generate multi-human motion beyond fixed-sized groups. To ensure fair comparison, we retrain several representative open-source baseline methods on our dataset and motion representation and compare them with our approach.

Notably, our method significantly outperforms on metrics such as FID, Diversity, and MModality. However, the improvements in R Precision and MM Dist are less pronounced. We attribute this to the evaluation methodology adopted from~\cite{guo2022generating}, which is tailored for text-to-motion scenarios involving one or two individuals. This limitation poses challenges for effectively training and assessing models designed to handle varying group sizes, particularly in terms of motion-language embedding alignment, which R Precision and MM Dist emphasize. Despite this, metrics like FID, Diversity, and MModality remain robust as they focus exclusively on motion embeddings. Qualitative results further highlight the superiority of our approach, demonstrating its effectiveness in generating high-quality, multi-human motions.

\subsection{Qualitative Results}
\begin{figure}[htbp]
  \centering
  \includegraphics[width=0.49\textwidth]{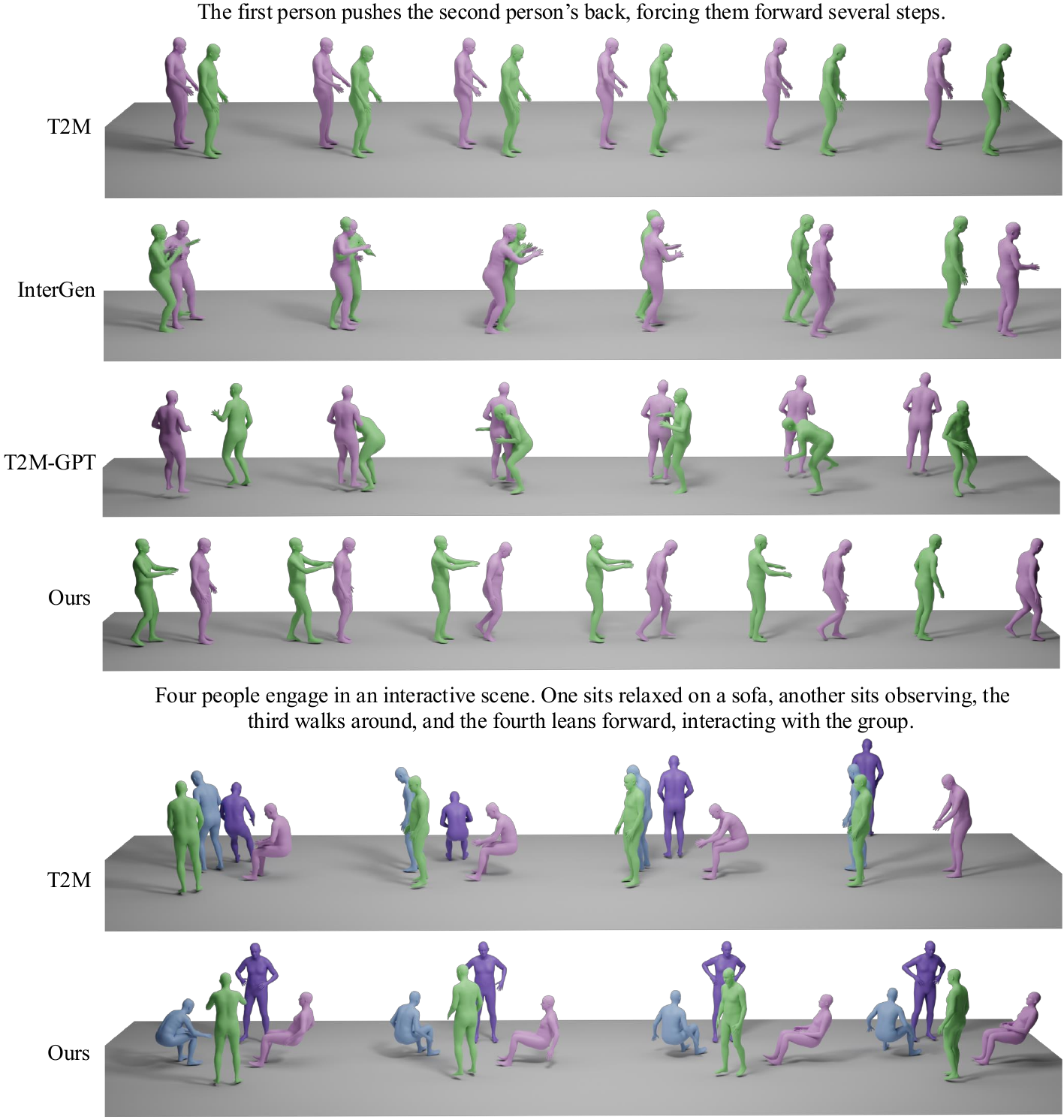}
  \caption{Text-to-motion Generation. Our method produces smoother, more accurate motions that align more precisely with textual descriptions compared to prior approaches.}
  \label{fig:t2m}
  \vspace{-0.3cm}
\end{figure}

We compare text-to-motion generation results between our method and previous approaches in Figure~\ref{fig:t2m}. The results show that motions generated by other methods are less aligned with textual descriptions and sometimes exhibit penetration issues (see InterGen results). In contrast, our method produces smoother and more accurate motions across different numbers of individuals. It effectively captures nuanced behaviors and spatial relationships, faithfully reflecting the given prompts.

\subsection{Other Motion-relevant Tasks}
With a shared space for motion and language tokens, our method naturally supports various motion-related tasks, demonstrating its versatility.

\begin{figure}[htbp]
  \centering
  \includegraphics[width=0.49\textwidth]{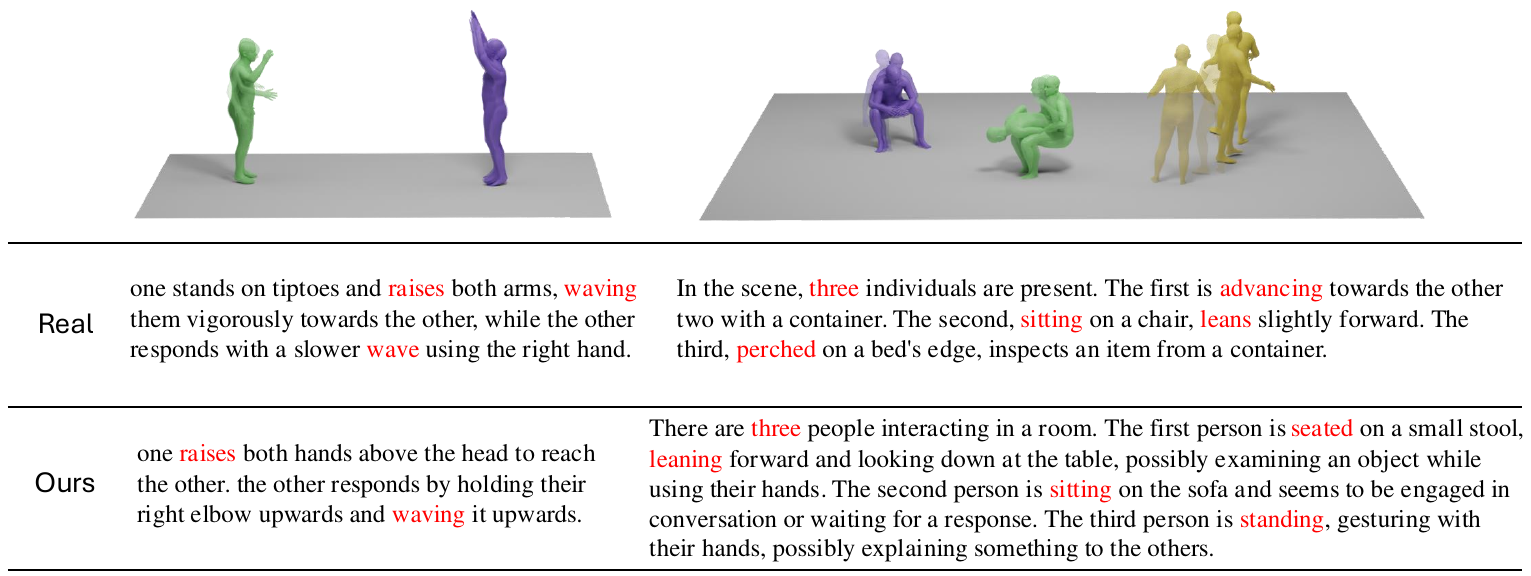}
  \caption{Motion to Text Generation. 'Real' is the labeled caption in the dataset. Our model effectively captures key actions and generates meaningful descriptions.}
  \label{fig:m2t}
\end{figure}

\noindent\textbf{Motion Captioning} 
Motion captioning generates textual descriptions from input. Figure~\ref{fig:m2t}  demonstrates that our model effectively captures key actions in social scenarios.

\begin{figure}[htbp]
  \centering
  \includegraphics[width=0.49\textwidth]{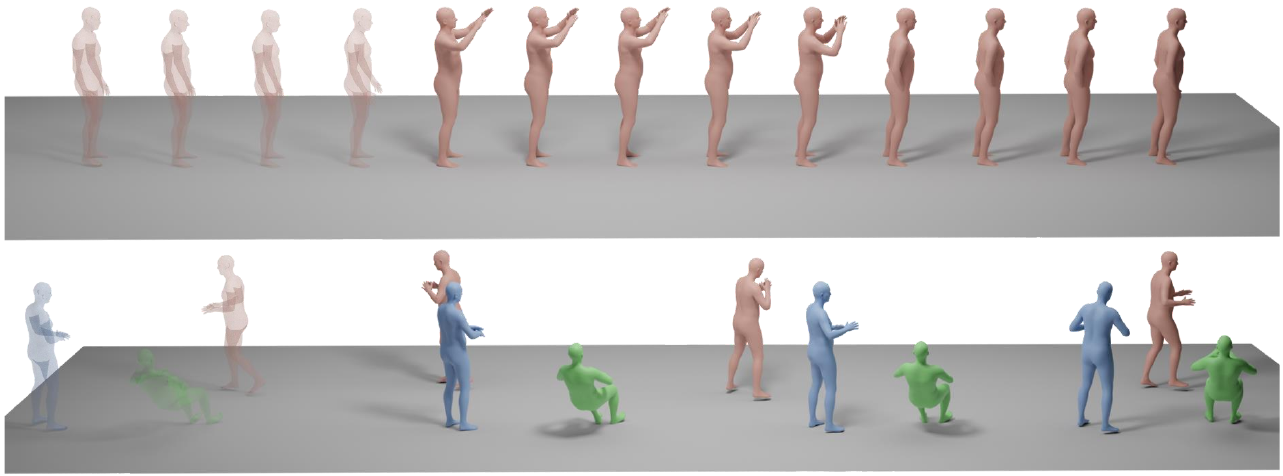}
  \caption{Motion Forecasting. Transparency indicates the initial motion used as a condition. Our model supports any number of individuals.}
  \label{fig:pred}
  \vspace{-0.3cm}
\end{figure}

\begin{figure}[htbp]
  \centering
  \includegraphics[width=0.49\textwidth]{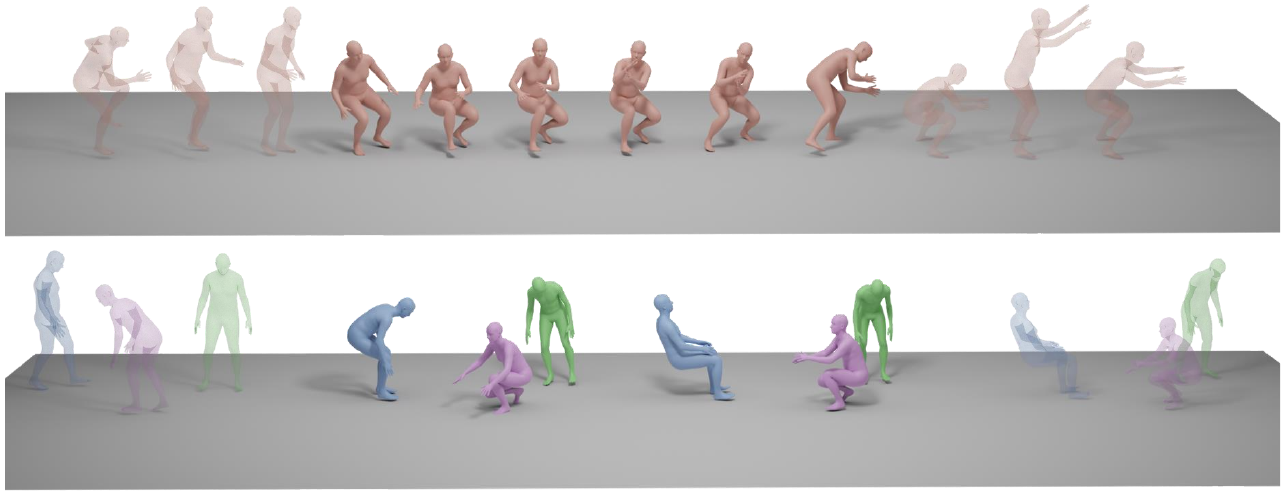}
  \caption{Motion Completion. Transparency indicates the initial and ending motions used as conditions.}
  \label{fig:inbetween}
  \vspace{-0.3cm}
\end{figure}

\noindent\textbf{Motion Completion} 
The Motion Completion task involves filling in missing motion segments based on given parts. 
Given only the start, it becomes motion forecasting (Figure~\ref{fig:pred}), while providing both start and end forms in-between completion (Figure~\ref{fig:inbetween}). The results demonstrate that our model effectively handles motion completion for varying numbers of individuals.

\begin{figure}[htbp]
  \centering
  \includegraphics[width=0.49\textwidth]{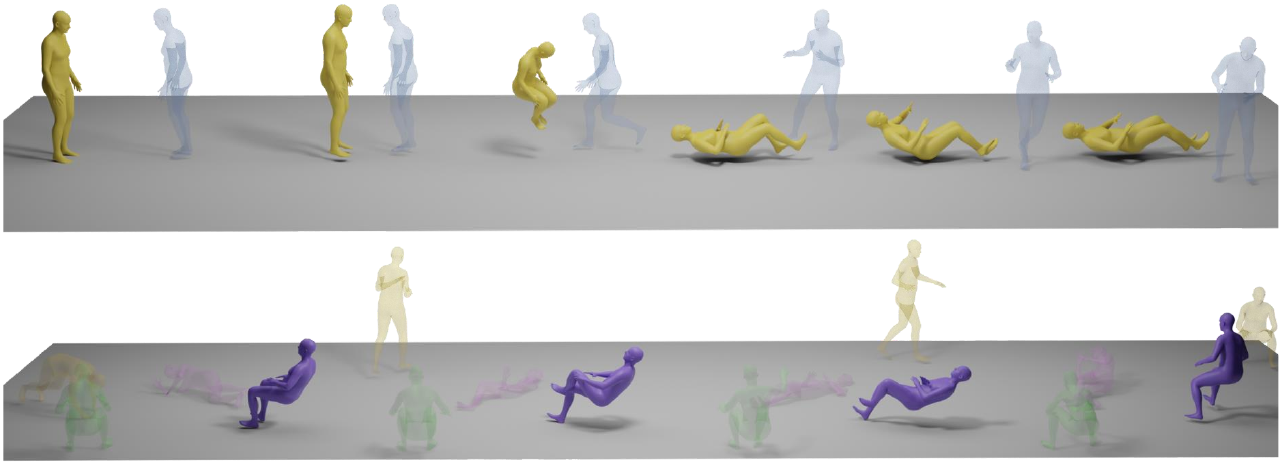}
  \caption{Reaction Generation. Transparency indicates the motions of the persons used as conditions. Our model accommodates any number of individuals as input conditions.}
  \label{fig:react}
  \vspace{-0.3cm}
\end{figure}

\noindent\textbf{Reaction Generation} 
Reaction Generation is a multi-person motion task that involves generating reaction motions based on existing motions. Unlike two-person scenarios, where one motion is used to generate another, our method can generate reactions for any number of individuals, as shown in Figure~\ref{fig:react}.

\subsection{Ablation Study}
\noindent\textbf{Motion Tokenizer Codebook Size} 
We explore the impact of varying codebook sizes for VQVAE-based tokenization on our SocialX dataset, with the results presented in Table~\ref{tb:token_abla}. The results indicate that increasing the codebook size improves the VQVAE's ability to reconstruct motion. After balancing reconstruction performance and codebook size, we select 512 as the codebook size in our experiment.


\begin{table}[h]
\centering
{\small
\renewcommand{\arraystretch}{.38}
\begin{tabular}{lcccc}
\toprule
\textbf{Dataset} & \textbf{\#Token} & \textbf{MPJPE↓} & \textbf{PAMPJPE↓} & \textbf{ACCEL↓} \\
\midrule
\multirow{4}{*}{SocialX} 
    & 128 & 80.01 & 48.40 & 7.41 \\
    & 256   & 82.17 & 45.60 & 7.27 \\
    & 512   & 67.25  & 38.82 & 6.86 \\
    & 1024   & 65.26  & 35.81 & 6.97 \\
\bottomrule
\end{tabular}
}
\caption{Ablation on VQVAE codebook size.The codebook size impacts model learning—a smaller size increases errors, while a larger size adds learning complexity.}
\label{tb:token_abla}
\end{table}


\noindent\textbf{Data Augmentation}
To enhance the model's robustness, we incorporate a data augmentation strategy during both the pre-training and post-training stages. This strategy involves randomly shuffling the order of humans in interactions, ensuring that the model's understanding of interactions remains invariant to participant order. 
This random shuffling not only makes the reference human for computing relative poses variable but also randomizes the order of the remaining participants. Consequently, the model learns to capture interaction dynamics independent of ordering, improving its generalization to diverse scenarios.
To assess the effectiveness of this augmentation, we conduct an ablation study. As shown in Table~\ref{tab:abla_aug}, this strategy significantly enhances the diversity of the generated outputs.

\begin{table}[h]
\centering
{\small
\renewcommand{\arraystretch}{.38}
\begin{tabular}{cccc}
\toprule
\textbf{Method} & \textbf{FID$\downarrow$} & \textbf{Diversity$\rightarrow$} & \textbf{MModality$\uparrow$} \\ 
\midrule
 Ground Truth  & 0.0002\textsuperscript{$\pm$.0001} & 14.334\textsuperscript{$\pm$.114} & - \\
Ours w/o Aug  & 1.794\textsuperscript{$\pm$.007} & 13.817\textsuperscript{$\pm$.136} & 3.141\textsuperscript{$\pm$.105} \\
 Ours & \textbf{1.403}\textsuperscript{$\pm$.005} & \textbf{14.213}\textsuperscript{$\pm$.147} & \textbf{3.263}\textsuperscript{$\pm$.098} \\ 
\bottomrule
\end{tabular}%
}
\caption{Ablation on data augmentation. The augmentation strategy of random shuffling enhances the model's interaction learning.}
\label{tab:abla_aug}
\vspace{-0.3cm}
\end{table}


\section{Conclusion and Discussion}
In this paper, we present \modelname, a unified framework with novel motion representations for multi-human scenarios, enabling social interaction generation, and other motion-language tasks. Further, we introduce SocialX, the first large-scale dataset for multi-human social interactions, and demonstrate significant improvements in motion generation through our benchmark.
However, challenges remain. Current metrics, such as R Precision and MM Dist, are inadequate for evaluating multi-human interactions, highlighting the need for improved evaluation methods. Additionally, extending models to open-domain scenarios with complex interactions and multimodal contexts is a key direction for future work. Our framework and benchmark drive motion-language research, fostering further exploration of scalable multi-human interaction modeling.


\section*{Acknowledgments}
This project was partially supported by NIH grant R01AG089169, Massachusetts AI \& Technology Center (MassAITC) P30AG073107, and the Stanford University Jaswa Innovator Awards. The authors would also
like to thank Chaitanya Patel, Mengyi Shan and Bowei Chen for their valuable input and discussions.

{
    \small
    \bibliographystyle{ieeenat_fullname}
    \bibliography{main}
}

\clearpage
\section{Appendix}

\begin{table*}[htp]
\centering
{\small
\renewcommand{\arraystretch}{.7}
\begin{tabular}{lccccccc}
\toprule
\textbf{Method}  & \textbf{\# Human} & \textbf{Generation} & \textbf{Captioning} & \textbf{Q\&A} & \textbf{Forecasting} &\textbf{Completion} & \textbf{Reaction Synthesis} \\ 
\midrule
MoMask~\cite{guo2024momask} & 1 & \cmark & \xmark   & \xmark  & \xmark & \xmark & \xmark \\
MotionGPT~\cite{jiang2023motiongpt}    & 1   & \cmark   & \cmark  & \cmark    & \cmark  & \cmark & \xmark \\
CoMA~\cite{sun2024coma} & 1 & \cmark & \cmark   & \cmark  & \cmark & \cmark & \xmark \\
ReGenNet~\cite{xu2024regennet} & 2    & \xmark  & \xmark & \xmark   & \xmark & \xmark & \cmark \\
ReMoS~\cite{ghosh2024remos} & 2    & \xmark  & \xmark & \xmark   & \xmark & \xmark & \cmark \\
InterGen~\cite{liang2024intergen}    & 2    & \cmark  & \xmark & \xmark   & \xmark & \xmark & \cmark \\
in2IN~\cite{ruiz2024in2in} & 2    & \cmark  & \xmark & \xmark   & \xmark  & \xmark & \cmark \\
InterMask~\cite{javed2024intermask}  & 2    & \cmark  & \xmark & \xmark   & \cmark & \cmark & \cmark \\
\cite{shan2024towards}  & abitrary    & \cmark   & \xmark & \xmark  & \xmark & \xmark & \xmark \\
FreeMotion~\cite{fan2024freemotion}  & abitrary    & \cmark   & \xmark & \xmark  & \xmark & \xmark & \cmark \\
\midrule
Ours & arbitrary    & \cmark    & \cmark  & \cmark  & \cmark & \cmark & \cmark \\
\bottomrule
\end{tabular}
}
\caption{Summary of various methods based on tasks. Green checkmarks (\cmark) indicate availability, and red crosses (\xmark) indicate unavailability. Our model not only handles an arbitrary number of people as input and output but is also capable of both generation and understanding social interactions, supporting an array of tasks including captioning, forecasting, reaction generation, and etc.}
\label{tab:review}
\end{table*}

\subsection{Pre-training and Post-training Prompts}
The prompt examples used in the pre-training and post-training stages are detailed in Table~\ref{tab:pre_prompt} and Table~\ref{tab:post_prompt}, respectively.


\begin{table*}
\centering
\caption{Prompt templates in our pre-training stage.}
\label{tab:pre_prompt}
\resizebox{0.7\textwidth}{!}{%
\begin{tabular}{lll}
\hline
\textbf{Task}                  & \textbf{Input}                                                                                     & \textbf{Output} \\ \hline
\textbf{Text-to-Motion}        & \begin{tabular}[c]{@{}l@{}} \texttt{<Caption\_Placeholder>} \end{tabular} & \texttt{<Motion\_Placeholder>} \\ \hline
\textbf{Motion-to-Text} & \begin{tabular}[c]{@{}l@{}} \texttt{<Motion\_Placeholder>} \end{tabular} & \texttt{<Caption\_Placeholder>} \\ \hline
\textbf{Motion Prediction}        & \begin{tabular}[c]{@{}l@{}}Predict motion: \texttt{<Motion\_part1>} \end{tabular} & \texttt{<Motion\_part2>} \\ \hline
\textbf{Reaction Generation}        & \begin{tabular}[c]{@{}l@{}}Generate Reaction: \texttt{<Motion\_react1>} \end{tabular} & \texttt{<Motion\_react2>} \\ \hline
\textbf{Motion Inbetween}        & \begin{tabular}[c]{@{}l@{}}Complete the masked motion: \texttt{<Motion\_Masked>} \end{tabular} & \texttt{<Motion>} \\ \hline
\end{tabular}%
}
\end{table*}

\begin{table*}
\centering
\caption{Examples of prompt templates in our post-training stage.}
\label{tab:post_prompt}
\resizebox{\textwidth}{!}{%
\begin{tabular}{lll}
\hline
\textbf{Task}                  & \textbf{Input}                                                                                     & \textbf{Output} \\ \hline
\textbf{Text-to-Motion}        & \begin{tabular}[c]{@{}l@{}}Show me a motion that captures the essence of <Caption>. \\ Can you generate a motion that represents the <Caption>? \end{tabular} & \texttt{<Motion>} \\ \hline
\textbf{Text-to-Motion w/ Frame Length} & \begin{tabular}[c]{@{}l@{}}I need a motion that lasts approximately <Frame> frames for the caption: <Caption>. \\ Can you create a motion sequence that lasts for <Frame> frames and represents <Caption> in motion?\end{tabular} & \texttt{<Motion>} \\ \hline
\textbf{Text-to-Motion w/ Second Length} & \begin{tabular}[c]{@{}l@{}}I need a motion that lasts <Second> seconds and conveys the message of <Caption>. \\ Can you create a motion that lasts <Second> seconds and demonstrates the concept of <Caption>?\end{tabular} & \texttt{<Motion>} \\ \hline
\textbf{Text-to-Motion w/ Human Number} & \begin{tabular}[c]{@{}l@{}}Please create a motion involving <Human> humans and illustrating the idea of <Caption>. \\Can you demonstrate <Caption> with a motion that includes <Human> humans?\end{tabular} & \texttt{<Motion>} \\ \hline
\textbf{FrameLength-to-Motion}      & \begin{tabular}[c]{@{}l@{}}Show me a motion that lasts for no more than <Frame> frames. \\ Can you make a motion that is shorter than <Frame> frames in length?\end{tabular} & \texttt{<Motion>} \\ \hline
\textbf{SecondLength-to-Motion}      & \begin{tabular}[c]{@{}l@{}}Give me a motion that has a length of <Second> seconds or less. \\ Can you make a motion that is no longer than <Second> seconds in duration?\end{tabular} & \texttt{<Motion>} \\ \hline
\textbf{HumanNumber-to-Motion}      & \begin{tabular}[c]{@{}l@{}}Create a motion that showcases <Human> humans performing unique activities. \\ Can you design a motion for <Human> humans that feels lifelike?\end{tabular} & \texttt{<Motion>} \\ \hline
\textbf{Random Motion}         & \begin{tabular}[c]{@{}l@{}}Create movements that are not anticipated. \\ Produce movements that are natural and unforced.\end{tabular} & \texttt{<Motion>} \\ \hline
\textbf{Motion-to-Text}        & \begin{tabular}[c]{@{}l@{}}What kind of motion is displayed in <Motion>? Describe it in text? \\ Describe the motion portrayed in <Motion> using words.\end{tabular} & \texttt{<Caption>} \\ \hline
\textbf{Motion-to-Text w/ Frame Length} & \begin{tabular}[c]{@{}l@{}}What is happening in <Motion> during a duration of <Frame> frames? \\ Describe the motion depicted in <Motion> over <Frame> frames.\end{tabular} & \texttt{<Caption>} \\ \hline
\textbf{Motion-to-Text w/ Second Length} & \begin{tabular}[c]{@{}l@{}}What is the action being demonstrated in <Motion> over <Second> seconds? \\ What is being demonstrated in <Motion> that is <Second> seconds long?\end{tabular} & \texttt{<Caption>} \\ \hline
\textbf{Motion-to-Text w/ Human Number} & \begin{tabular}[c]{@{}l@{}}What is happening among <Human> humans in <Motion>? \\ Describe the coordinated actions of <Human> humans in <Motion>.\end{tabular} & \texttt{<Caption>} \\ \hline
\textbf{Motion-to-FrameLength}      & \begin{tabular}[c]{@{}l@{}}What is the duration of <Motion>'s gestures in frames? \\ Compute the frame count for <Motion>'s body movements?\end{tabular} & \begin{tabular}[c]{@{}l@{}}There are <Frame> frames in the motion. \\ The length of given motion is about <Frame> frames.\end{tabular} \\ \hline
\textbf{Motion-to-SecondLength}      & \begin{tabular}[c]{@{}l@{}}How many seconds are there in <Motion>? \\ Calculate the second duration for <Motion>'s actions.\end{tabular} & \begin{tabular}[c]{@{}l@{}}There are about <Second> seconds in the motion. \\ The motion lasts for roughly estimated <Second> seconds.\end{tabular} \\ \hline
\textbf{Motion-to-HumanNumber}      & \begin{tabular}[c]{@{}l@{}}How many people are shown in <Motion>? \\ Determine the number of individuals involved in <Motion>.\end{tabular} & \begin{tabular}[c]{@{}l@{}} A total of <Human> individuals are participating. \\ The scene includes <Human> humans. \end{tabular} \\ \hline
\textbf{Caption-to-FrameLength}     & \begin{tabular}[c]{@{}l@{}}Predict the frame count required for the motion corresponding to <Caption>. \\ How many frames should the motion that matches <Caption> have?\end{tabular} & \begin{tabular}[c]{@{}l@{}}The motion has an estimated duration of <Frame> frames. \\ The total number of frames in the motion is roughly <Frame>.\end{tabular} \\ \hline
\textbf{Caption-to-SecondLength}     & \begin{tabular}[c]{@{}l@{}}Estimate the expected number of seconds required for the motion that matches <Caption>. \\ What is the expected second length for the motion that corresponds to <Caption>?\end{tabular} & \begin{tabular}[c]{@{}l@{}}The motion's second count is <Second>. \\ The motion's second count is roughly estimated to be <Second>.\end{tabular} \\ \hline
\textbf{Caption-to-HumanNumber}     & \begin{tabular}[c]{@{}l@{}}How many humans are involved in the motion described by <Caption>? \\ Determine the number of individuals participating in <Caption>.\end{tabular} & \begin{tabular}[c]{@{}l@{}}The motion involves <Human> humans. \\ There are <Human> people in this motion.\end{tabular} \\ \hline
\textbf{FrameLength-to-Caption}     & \begin{tabular}[c]{@{}l@{}}Based on the <Frame> frames of the motion, what is the likelihood of it being a full-body movement or a partial-body movement? \\ Based on the motion length <Frame> frames, what is the likelihood of it being a cardiovascular or respiratory exercise? \end{tabular} & \texttt{<Caption>} \\ \hline
\textbf{SecondLength-to-Caption}     & \begin{tabular}[c]{@{}l@{}}Given <Second> seconds of motion, what body parts are likely to be involved? \\ Predict the type of sport or exercise that would require <Second> seconds of motion. \end{tabular} & \texttt{<Caption>} \\ \hline
\textbf{HumanNumber-to-Caption}     & \begin{tabular}[c]{@{}l@{}}Generate a description of the motion involving <Human> humans. \\ What types of activities could <Human> humans perform together? \end{tabular} & \texttt{<Caption>} \\ \hline
\textbf{Random Caption}        & \begin{tabular}[c]{@{}l@{}}Write a brief summary of how someone might move their feet while doing the foxtrot. \\ Give me a motion description.\end{tabular} & \texttt{<Caption>} \\ \hline
\textbf{Motion Prediction}        & \begin{tabular}[c]{@{}l@{}}Predict motion: <Motion\_part1> \\ Do the motion prediction task for <Motion\_part1>.\end{tabular} & \texttt{<Motion\_part2>} \\ \hline
\textbf{Reaction Generation}        & \begin{tabular}[c]{@{}l@{}}Generate Reaction: <Motion\_react1> \\ Do the reaction generation task for <Motion\_react1>.\end{tabular} & \texttt{<Motion\_react2>} \\ \hline
\textbf{Motion Inbetween}        & \begin{tabular}[c]{@{}l@{}}Complete the masked motion: <Motion\_Masked> \\ Here is a masked motion sequence <Motion\_Masked>, complete it.\end{tabular} & \texttt{<Motion>} \\ \hline
\end{tabular}%
}
\end{table*}

\subsection{Prompts for Motion Captioning}
To generate captions for motion sequences using the corresponding videos, we leverage GPT-4o-mini~\cite{openai2024gpt4o} with a two-stage captioning process. The first stage provides an overall description of the entire scene, while the second stage refines these captions to focus specifically on human motions, excluding human-object interactions, appearance, and other unrelated information. The prompts for these two stages are detailed below:

\noindent\textbf{Stage 1: Overall Scene Captioning}
\begin{itemize}
    \item You are an AI visual assistant equipped to analyze video content.
    \item You will be shown frames from different camera views of the same scene. The frames are uniformly sampled from multi-human, multi-object interaction videos. Multi-view videos are captured from different cameras in the same scene to help you better understand human interactions.
    \item Please describe the scene by:
    \begin{itemize}
        \item Providing a general description of how many people are present and their collective actions.
        \item Giving detailed descriptions of each person's actions using the format: \textit{The first person is ..., the second person is ..., etc.}
    \end{itemize}
    \item Note: Combine all the frames of the same motion video and the given general description as if observing the entire video.
    \item The descriptions must follow these rules:
    \begin{itemize}
        \item Do not describe the appearance of individuals; refer to them as \textit{the first person}, \textit{the second person}, etc. Avoid mentioning camera IDs.
        \item Focus on atomic actions of body parts and describe their temporal sequence.
        \item Keep the description under 100 words, avoiding redundancy.
        \item Write the description as a single paragraph without line breaks or special symbols.
    \end{itemize}
\end{itemize}

\noindent\textbf{Stage 2: Motion-Focused Refinement}
\begin{itemize}
    \item You are an AI assistant specialized in analyzing and refining descriptions of multi-human interaction motions.
    \item Your task is to focus exclusively on describing human motions while ensuring accuracy.
    \item The descriptions may include:
    \begin{itemize}
        \item Human-object interactions.
        \item Human-human interactions.
        \item Human-human-object interactions.
    \end{itemize}
    \item Refine the description to:
    \begin{itemize}
        \item Focus solely on the motion and interaction dynamics of the humans involved.
        \item Retain accurate information about the number of humans present in the interaction and video.
        \item Exclude irrelevant details about the appearance of people or objects, unless they directly influence the motion.
        \item Ensure the description is concise, clear, and free from ambiguity.
    \end{itemize}
\end{itemize}

\subsection{Statistics}
\begin{figure}[htp]
  \centering
  \includegraphics[width=0.47\textwidth]{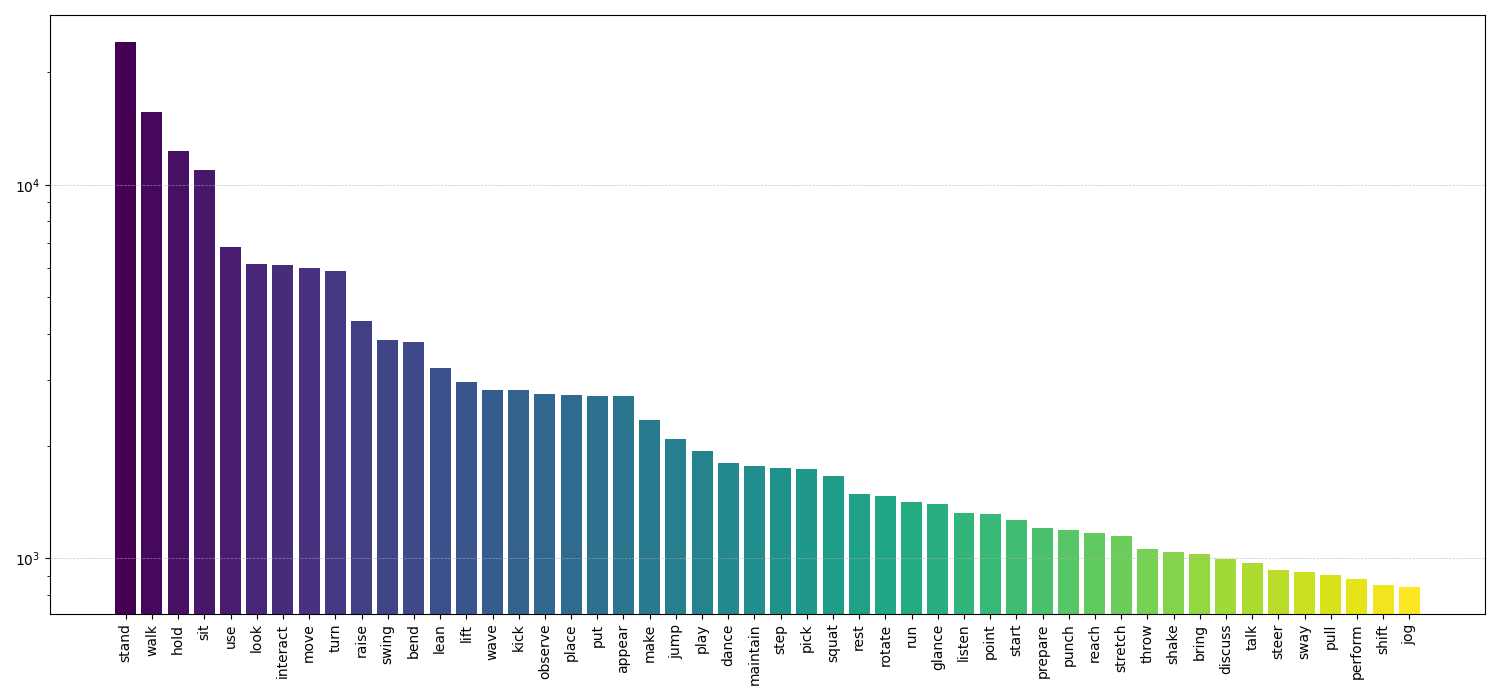}
  \caption{Number of occurrences of each motion type in our SocialX dataset.}
  \label{motion_type}
\end{figure}

The SocialX dataset includes a diverse collection of over 50 motion types, each accompanied by detailed language descriptions to ensure clarity and usability. A summary of the motion types is presented in Figures~\ref{motion_type}.

\begin{figure}[htp]
  \centering
  \includegraphics[width=0.47\textwidth]{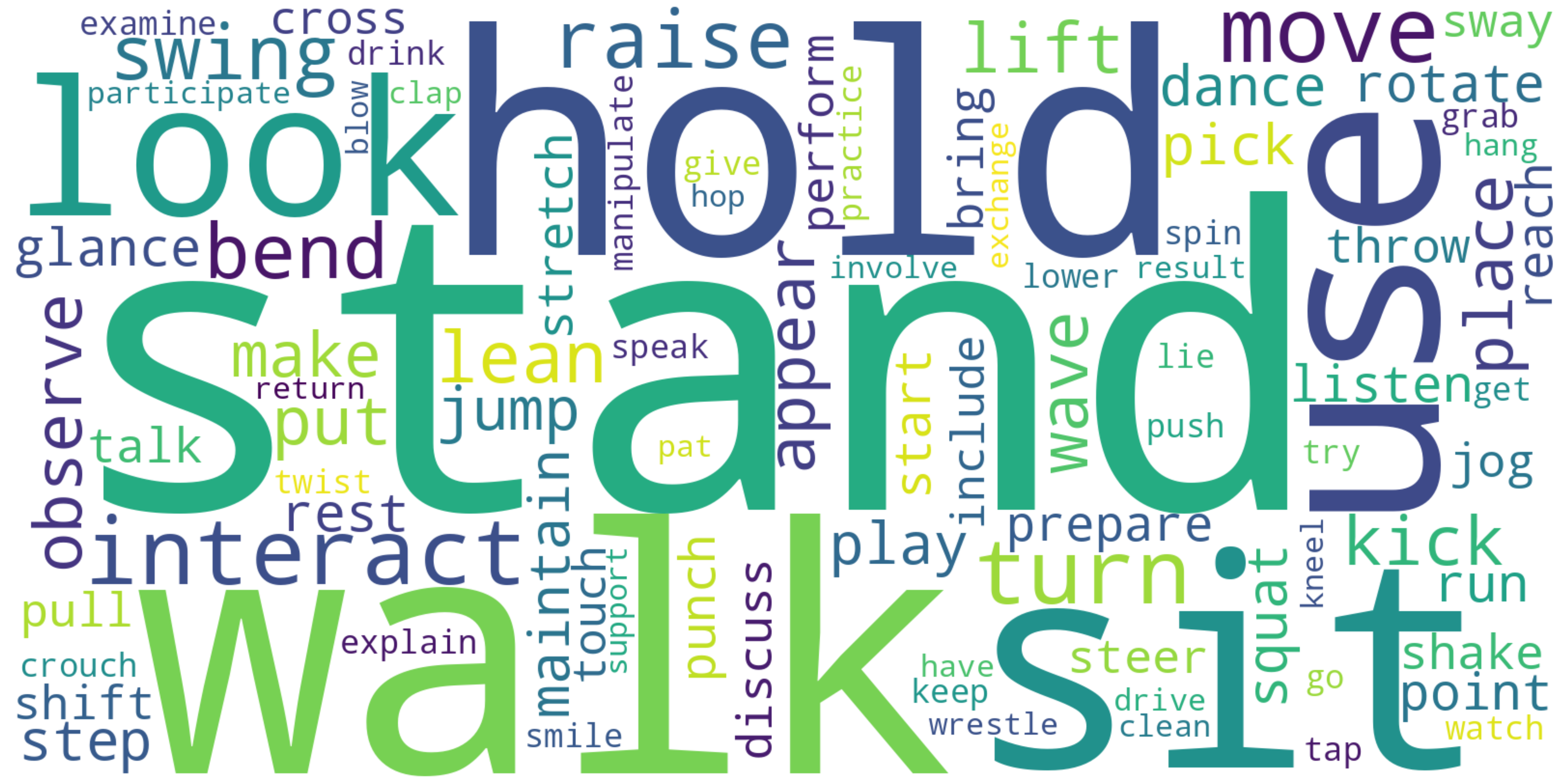}
  \caption{Word cloud built from the text descriptions in our SocialX dataset.
  }
  \label{word_cloud}
\end{figure}

A word cloud of the associated descriptions for Figures~\ref{motion_type} is presented in \ref{word_cloud}.

\begin{table}[htp]
\centering
\resizebox{0.48\textwidth}{!}{%
\begin{tabular}{lccccccc}
\toprule
\textbf{Dataset}  & \textbf{\#Seq} & \textbf{\#Frm} & \textbf{Repr} & \textbf{Desp}\\ 
\midrule
HumanML3D~\cite{guo2022generating} & 14,616 & 2M & H3D   & \cmark  \\ 
InterHuman~\cite{liang2024intergen} & 7,779 & 1.7M & IH   & \cmark  \\ 
InterX~\cite{xu2024inter} & 11,388 & 8.1M & IX   & \cmark  \\ 
HOI-M$^3$~\cite{zhang2024hoi} & 7245 & 18.1M & SMPLX   & \xmark  \\ 
MuPoTs-3D~\cite{mehta2018single} & 20 & 8K & Kpt3D & \xmark  \\ 
Panoptic Haggling~\cite{joo2019towards} & 150 & 850K & Kpt3D  & \xmark  \\ 
\midrule
SocialX (ours) & \textgreater 40K & \textgreater 30M &  XH3D & \cmark  \\
\bottomrule
\end{tabular}
}
\caption{Summary of dataset statistics.}
\label{data_sum}
\end{table}

\subsection{Dataset Summary.}
We list the information of datasets we leverage in Table~\ref{data_sum} and compare them with our SocialX.\\
\noindent\textbf{HumanML3D} HumanML3D dataset~\cite{guo2022generating} is a single-human motion dataset comprising 14,616 motion sequences sourced from AMASS~\cite{mahmood2019amass} and HumanAct12~\cite{guo2020action2motion}, accompanied by 44,970 sequence-level textual descriptions. \\
\noindent\textbf{InterHuman} InterHuman dataset~\cite{liang2024intergen} is a two-human interaction dataset containing 7,779 interaction sequences with 23,337 textual descriptions. \\
\noindent\textbf{InterX} InterX dataset~\cite{xu2024inter} is a two-human interaction dataset containing  11,388 interaction sequences, each paired with 3 distinct textual annotations. \\
\noindent\textbf{HOI-M$^3$} HOI-M$^3$ dataset~\cite{zhang2024hoi} is a multiple
human-object interaction dataset within different contextual environments. It includes
181 million frames featuring 46 subjects engaged in interactions with 90 objects.\\
\noindent\textbf{MuPoTs-3D} MuPoTS-3D (multi-human Pose Estimation Test Set in 3D)~\cite{mehta2018single} is a dataset designed for pose estimation, consisting of over 8,000 frames captured from 20 real-world scenes featuring up to three subjects. Each pose is annotated using a 14-point skeleton model. \\
\noindent\textbf{CMU Panoptic} The CMU Panoptic dataset~\cite{joo2019towards} provides 3D pose annotations for multi-human social activities across 65 videos (5.5 hours) with multi-view annotations. Of these, only 17 feature multi-human scenarios with accurate camera parameters. We use the Haggling sequences, as they include multiple individuals and provide reliable pose annotations suitable for our framework.

\subsection{Implementation Details}
We configure the motion tokenizer with a codebook size of $K = 512$ for our experiments. The motion encoder adopts a temporal downsampling rate of 4, following~\cite{jiang2023motiongpt}, to ensure efficient processing while maintaining motion fidelity. For the language model backbone, we utilize the T5-base architecture~\cite{raffel2020exploring}, consisting of 12 layers in both the transformer encoder and decoder. The feed-forward networks have a hidden dimensionality of $d_\text{ff} = 3072$, and the attention mechanism operates with an inner dimensionality of $d_\text{kv} = 64$. During this stage, a batch size of 256 is employed. 

For optimization, we employ the AdamW optimizer~\cite{loshchilov2017decoupled}, with an initial learning rate of $2 \times 10^{-4}$ for the pre-training stage and $1 \times 10^{-4}$ for the instruction tuning stage. The motion tokenizer is trained for 3K iterations, while the language model is trained for 1K iterations during the pre-training stage and an additional 1K iterations during the instruction tuning stage. Both stages utilize a batch size of 16. All experiments are conducted on 8 NVIDIA GeForce RTX 3090 GPUs using PyTorch framework~\cite{paszke2019pytorch}. 

\subsection{Evaluation Metrics}
To evaluate our model, we employ a range of metrics that comprehensively assess motion quality, diversity, multimodality, and text-motion alignment. For motion quality, we utilize Frechet Inception Distance (FID) \cite{heusel2017gans}, which measures the distance between the feature distributions of generated and real motions, providing a quantitative measure of fidelity. To evaluate diversity, we calculate the Diversity (DIV) metric, which assesses the variance in motion features across different generated samples, and Multimodality (MM), which quantifies the ability to generate multiple distinct motions conditioned on the same textual prompt. For text-motion alignment, we adopt R-Precision (Top-1, Top-2, Top-3) to measure how accurately the generated motions match their corresponding textual descriptions by ranking the Euclidean distances between motion and text embeddings. Additionally, we use Multi-Modal Distance (MM Dist) to compute the average Euclidean distance between motion embeddings and text embeddings, quantifying semantic coherence between motions and texts. We retrain the feature extractor using our SoicalX dataset following~\cite{guo2022generating}.

\subsection{Additional Qualitative Results}
\begin{figure*}[h]
  \centering
  \includegraphics[width=0.92\textwidth]{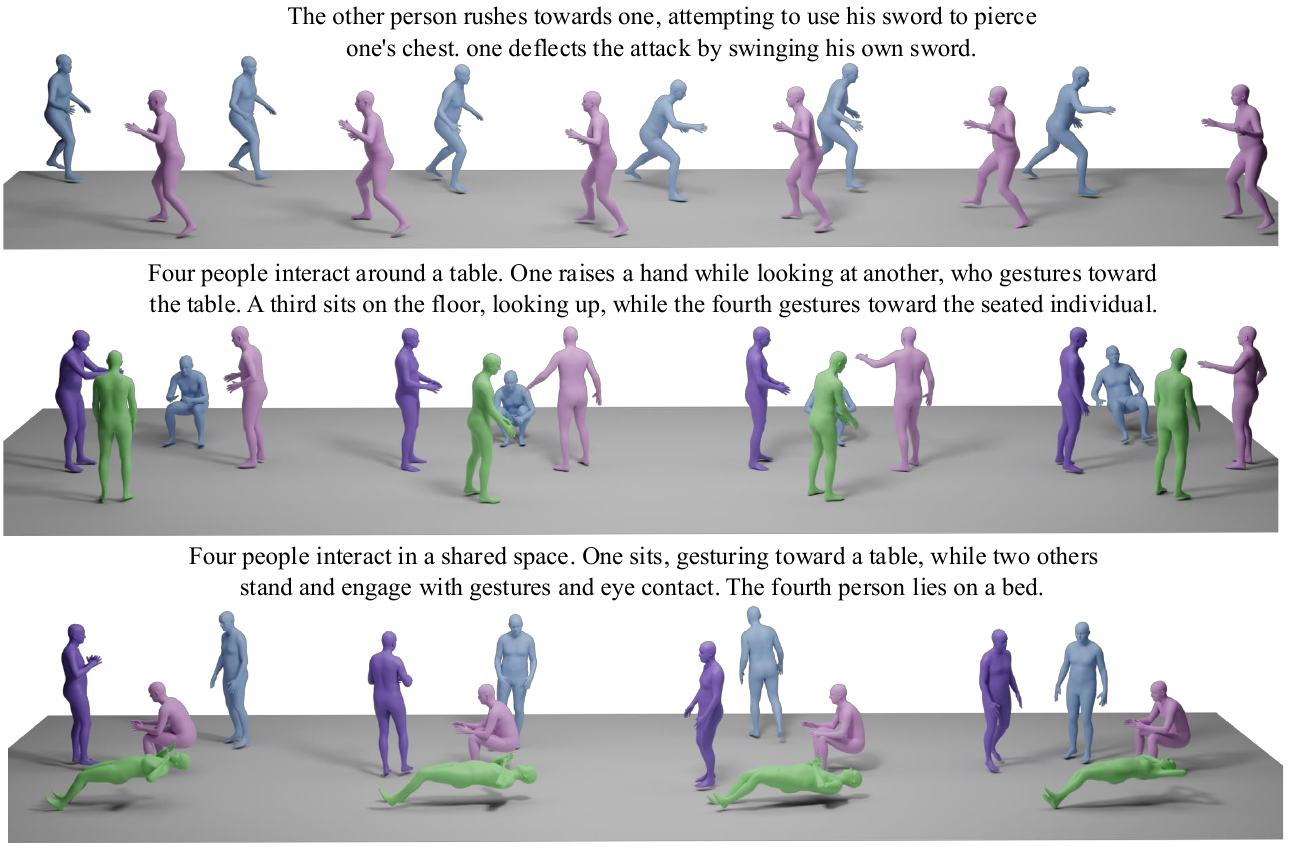}
  \caption{Additional results on Text to Motion Generation task.}
  \label{fig:t2m_app}
\end{figure*}

We show more qualitative results on the Text to Motion Generation task in Figure~\ref{fig:t2m_app}.

\end{document}